%% file: paper_cvpr.tex
\documentclass[10pt,twocolumn,letterpaper]{article}

\usepackage[pagenumbers]{cvpr} % To force page numbers, e.g. for an arXiv version

\makeatletter
\@namedef{ver@everyshi.sty}{}
\makeatother
\usepackage{tikz}

\usepackage{graphicx}
\usepackage{amsmath}
\usepackage{amssymb}
\usepackage{booktabs}
\usepackage{tabularx}
\usepackage{textcomp}
\usepackage[nolist, nohyperlinks, printonlyused]{acronym} % For consistent acronyms
\usepackage[colorinlistoftodos,prependcaption,textsize=tiny]{todonotes}
\usepackage{amssymb}% http://ctan.org/pkg/amssymb
\usepackage{pifont}% http://ctan.org/pkg/pifont

\newcommand{\cmark}{\ding{51}}%
\newcommand{\xmark}{\ding{55}}%

\usepackage[pagebackref,breaklinks,colorlinks]{hyperref}

\usepackage[capitalize]{cleveref}
\crefname{section}{Sec.}{Secs.}
\Crefname{section}{Section}{Sections}
\Crefname{table}{Table}{Tables}
\crefname{table}{Tab.}{Tabs.}

\def\cvprPaperID{11} % *** Enter the CVPR Paper ID here
\def\confName{CVPR}
\def\confYear{2023}

\begin{document}

\title{From Model-Based to Data-Driven Simulation:\\ Challenges and Trends in Autonomous Driving}

\author{Ferdinand Mütsch$^{1}$\textsuperscript{\textasteriskcentered}, Helen Gremmelmaier$^{2}$\textsuperscript{\textasteriskcentered}, Nicolas Becker$^{1}$\textsuperscript{\textasteriskcentered}\\
Daniel Bogdoll$^{2}$, Marc René Zofka$^{2}$, J. Marius Zöllner$^{1,2}$\\\\
$^{1}$KIT Karlsruhe Institute of Technology, Karlsruhe, Germany\\
$^{2}$FZI Research Center for Information Technology, Karlsruhe, Germany\\
{\tt\small muetsch@kit.edu},
{\tt\small gremmelmaier@fzi.de},
{\tt\small nicolas.becker@student.kit.edu},\\
{\tt\small bogdoll@fzi.de},
{\tt\small zofka@fzi.de},
{\tt\small zoellner@fzi.de}
}

% \author{Ferdinand Mütsch\\
% {\small KIT} \\
% {\tt\small muetsch@kit.edu}
% \and
% Helen Gremmelmaier\\
% {\small FZI}\\
% {\tt\small gremmelmaier@fzi.de}
% \and
% Nicolas Becker\\
% {\small KIT}\\
% {\tt\small nicolas.becker@student.kit.edu}
% \and
% Daniel Bogdoll\\
% {\small FZI}\\
% {\tt\small bogdoll@fzi.de}
% \and
% Marc René Zofka\\
% {\small FZI}\\
% {\tt\small zofka@fzi.de}
% \and
% J. Marius Zöllner\\
% {\small FZI}\\
% {\tt\small zoellner@fzi.de}
%}
\maketitle

\begingroup\renewcommand\thefootnote{\textasteriskcentered}
\footnotetext{These authors contributed equally}
\endgroup

%%%%%%%%% ABSTRACT
\begin{abstract}
\input{sections_paper/0_abstract}
\end{abstract}

%%%%%%%%% BODY TEXT
\input{sections_paper/1_introduction}
\input{sections_paper/2_levels}
\input{sections_paper/3_challenges_trends}
\input{sections_paper/4_conclusion}
\input{sections_paper/5_acknowledgment.tex} %Re-insert after review

%%%%%%%%% REFERENCES
{\small
\bibliographystyle{ieee_fullname}
\bibliography{references}
}

\end{document}

%% file: sections_paper/0_abstract.tex
Simulation is an integral part in the process of developing autonomous vehicles and advantageous for training, validation, and verification of driving functions. Even though simulations come with a series of benefits compared to real-world experiments, various challenges still prevent virtual testing from entirely replacing physical test-drives. Our work provides an overview of these challenges with regard to different aspects and types of simulation and subsumes current trends to overcome them. We cover aspects around perception-, behavior- and content-realism as well as general hurdles in the domain of simulation. Among others, we observe a trend of data-driven, generative approaches and high-fidelity data synthesis to increasingly replace model-based simulation.

% Vorschläge:
% Simulation in Autonomous Driving: From model to data driven approaches / aiming at data driven simulation
% From Model-Based to Data-Driven Simulation: Trends and Challenges in Autonomous Driving

%% file: sections_paper/1_introduction.tex
%%%%%%%%%%%%%%%%%%%%%%%%%%%%%%%%%%%%%%%%%%%
\section{Introduction}
\label{sec:introduction}
%%%%%%%%%%%%%%%%%%%%%%%%%%%%%%%%%%%%%%%%%%%

Simulations are an integral and indispensable part in the development process of autonomous vehicles (AV) and especially helpful for validation \& verification (V\&V). They enable researchers to massively scale up training and testing of their software stacks beyond the limits of real-world experiments. Moreover, they allow to better cover the "long tail of events". Particularly challenging corner case situations, or anomalies, only occur very rarely in reality, but are exceptionally powerful to improve safety and robustness of AVs. Simulation allows provoking them much more frequently. However, despite all their benefits, simulations still face a series of major challenges, especially related to an insufficiently large gap in realism compared to the real world. %We aim to outline these challenges and according trends.

\begin{figure}
    \includegraphics[width=83mm]{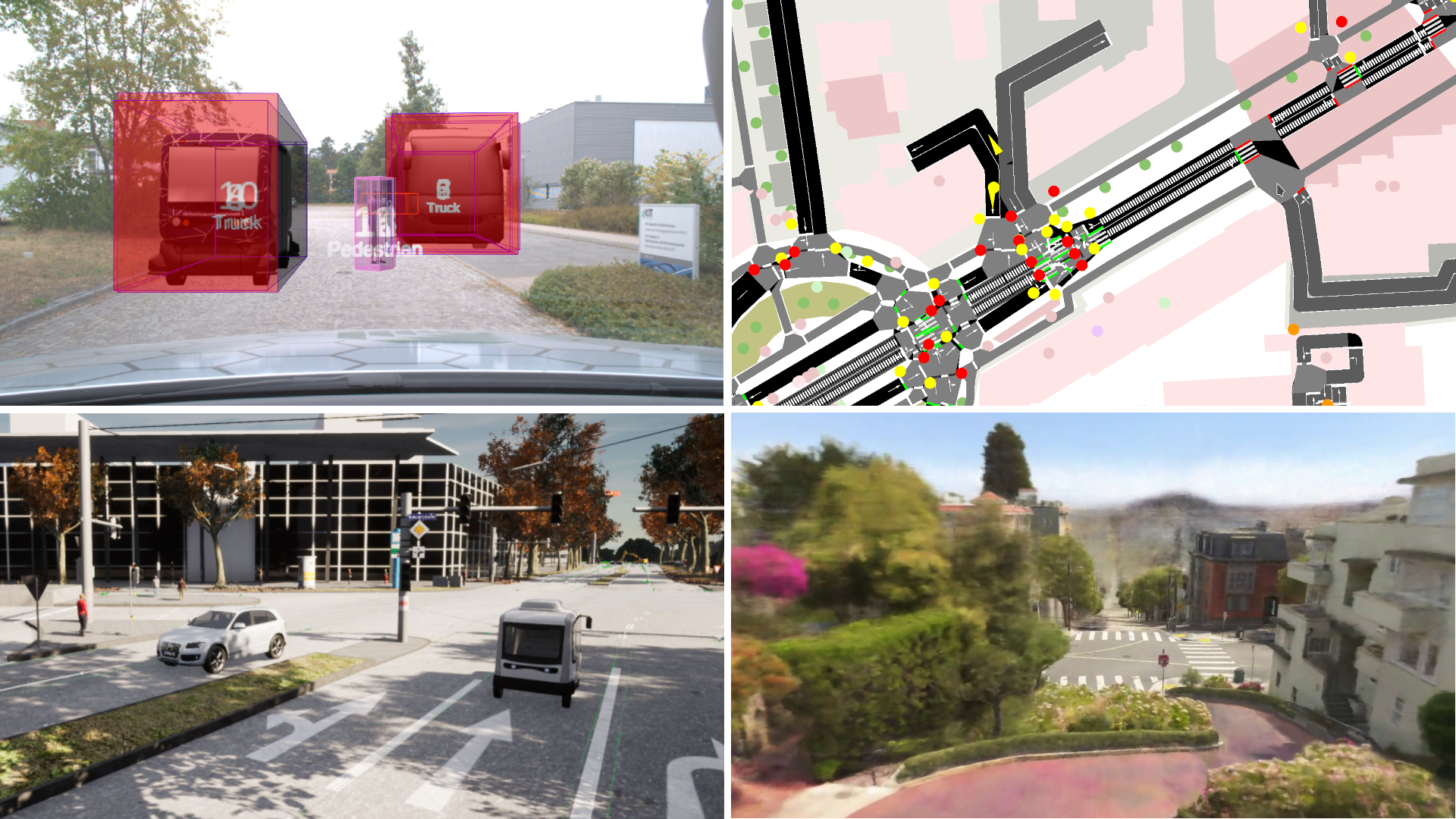}
    \centering
    \caption[Examples of our proposed simulation levels. Top-left to bottom-right: AR-enhanced (level 1), SUMO (level 2), CARLA (level 3), Block-NeRF (level 4)]{Examples of our proposed simulation levels. Top-left to bottom-right: AR-enhanced (level 1), SUMO (level 2), CARLA (level 3), Block-NeRF (level 4) ~\cite{waymo_block-nerf_website}.}
    \label{fig:simulator_screenshots}
\end{figure}

An increasing interest in driving simulation has brought forth detailed surveys about requirements, frameworks and their application in V\&V of AD recently. A technical report by Fadaie~\cite{Fadaie2019} and the studies from Kaur et al.~\cite{Kaur2021} and Kang et al.~\cite{Kang2019} aim at requirements analysis and comparison of existing simulation environments. Alghodhaifi and Lakshmanan \cite{Alghodhaifi2021} put
simulation methods and environments in the context of different assessment methods and focus on methodology. While these works provide rich overviews about the state-of-the-art, none of them specifically presents latest trends and their relevance for different challenges related to content-, behavior- or perception realism. During our literature review, we were missing clear indications about future developments in the field, especially in view of recent years' advancements towards highly data-driven methods. Also, to our knowledge, no up-to-date classification scheme exists, that could facilitate comparisons between simulation approaches. Our present work aims to fill this research gap.

In Sec.~\ref{sec:levels} we first establish a hierarchical classification of simulation approaches and touch upon their progression over time. Sec.~\ref{sec:challenges_trends} then surveys current challenges that simulations are confronted with, especially with respect to three major aspects of realism. Novel, trending concepts to overcome these challenges are showcased and give an indication about the direction future research may be moving towards. We focus on simulation for testing and refinement of AD software stacks in particular.

% Brauch man das? Kann das der Leser nicht selbst rausfinden oder kurz ins Inhaltsverzeichnis schauen?
% This paper is organized as follows: After this introduction in Sec.~\ref{sec:introduction}, Sec.~\ref{sec:levels} continues with our classification scheme for different types of simulation approaches by complexity and maturity. Sec.~\ref{sec:challenges_trends} depicts current shortcomings and challenges in simulation and how they are being addressed by recent trends. Eventually, Sec.~\ref{sec:conclusion} draws a conclusion and briefly touches upon relevant open research question on this topic.

%% file: sections_paper/2_levels.tex
%%%%%%%%%%%%%%%%%%%%%%%%%%%%%%%%%%%%%%%%%%%
\section{Levels of Simulation Approaches}
\label{sec:levels}
%%%%%%%%%%%%%%%%%%%%%%%%%%%%%%%%%%%%%%%%%%%

To better understand the context of different challenges and trends, we propose the classification scheme in Tab.~\ref{table:levels}. It delineates AD simulation approaches according to their comprehensiveness of covered aspects, their level of realism and other criteria. The presented hierarchy is partly inspired by \cite{NVIDIA_WAD}. Compared to previous taxonomies (\hspace{1sp}\cite{Fadaie2019, Kaur2021, Kang2019, Alghodhaifi2021}), the categories presented in our scheme are less informed by \textit{what} is simulated, but rather by \textit{how} things are simulated. That is, with a focus on the methodology used by a simulation. We compare the categories with respect to the following different aspects and additionally list typical applications and concrete examples of representatives of each category, where possible.

\begin{itemize}
    \itemsep0em 
    \item \textbf{Closed-loop / Reactive}:  This criterion describes whether these types of environments allow for closed-loop simulations, i.e. such in which the simulated system (or parts of it) can dynamically react to changed conditions in their environment and vice versa.
    \item \textbf{End-to-end (E2E) Development \& Testing}: This aspect is about whether development and testing of entire AD software stacks is supported by the simulation. Besides reactivity, the simulation environment must provide exteroceptive and interoceptive sensor models. Whether a given simulation environment is suitable for testing a particular AD stack, however, depends on the stack's specific requirements.
    \item \textbf{Visual Fidelity}: It describes the perceived degree of realism of observations produced by simulated sensors. This serves as a qualitative measure for how small the appearance gap is with the respective simulation type, i.e., relates to \emph{how things look like}.
    \item \textbf{Content \& Behavior Diversity}: This criterion is a qualitative measure for how much diversity a simulation allows for. This includes diversity in static and dynamic objects (e.g., different vehicle- or building shapes and textures), in the environment (e.g., different lighting conditions or weather) as well as in the behavior of dynamic agents within the simulation.
    \item \textbf{Object Representation}: This relates to the way in which static and dynamic objects are represented in the simulation. We distinguish between explicit and implicit representation. For explicitly represented objects, a precise description or model (e.g. a CAD model) is given, involving parameters that can be interpreted and tweaked purposefully. An implicit representation does not feature a clear-cut object description, but instead involves either sensor observations or a (latent) neural representation. The latter is usually not open to human interpretation. 
    \item \textbf{Scalability}: This criterion describes the degree at which new simulations can be designed and created at a large scale. It relates to how much manual effort is involved with the creation of new, diverse simulations. 
    \item \textbf{Controllability}: This criterion describes the degree of granularity at which parameters of a given simulation can be tweaked, that is, the level of control a user has over a simulation. While model-driven simulations allow to explicitly tweak variables such as a car's target speed, parameters are often times not even human-understandable with data-driven models.
\end{itemize}

\begin{flushleft}
\begin{table*}[]
\caption{Categories of Simulation Approaches for AD}
\scriptsize
\begin{tabularx}{\linewidth}{@{}>{\raggedright\arraybackslash}X|>{\raggedright\arraybackslash}X>{\raggedright\arraybackslash}X>{\raggedright\arraybackslash}X>{\raggedright\arraybackslash}X>{\raggedright\arraybackslash}X>{\raggedright\arraybackslash}X>{\raggedright\arraybackslash}X>{\raggedright\arraybackslash}X>{\raggedright\arraybackslash}X@{}}
\toprule
                                                                       & \textbf{Closed-loop / reactive} & \textbf{End-to-end development \& testing}            & \textbf{Visual fidelity}           & \textbf{Diversity (content \& behavior)} & \textbf{Object representation}                                     & \textbf{Scalability}          & \textbf{Control-lability}          & \textbf{Key use cases}                                       & \textbf{Examples}                                                                                                 \\ \midrule
{\textbf{0. Log replay}}                    & {\xmark}         & {\xmark} & {high}          & {low}                 & {implicit}               & {very low} & {none}          & {Perception}                   & {-}                                                                                              \\ \midrule
{\textbf{1. AR-enhanced log replay}}        & {partly}     & {\xmark} & {mixed}         & {medium}              & {mixed}             & {low}      & {low}           & {Perception}    & ~\cite{Zofka2015, Zofka2018, Koduri2018, Zhao2019}                                              \\ \midrule
{\textbf{2. Abstract dynamics simulation}} & {\cmark}        & {\xmark}   & {-}             & {medium}              & {explicit} & {medium}   & {high}          & {Prediction, planning, control} & {SUMO \cite{behrisch2011sumo}, CityFlow \cite{zhang_cityflow_2019}}                              \\ \midrule
{\textbf{3. Model-driven 3D simulation}}   & {\cmark}        & {\cmark}                              & {low - high}    & {medium}              & {explicit}               & {medium}   & {high}          & {E2E training / testing}           & {CARLA \cite{Dosovitskiy2017}, DriveSim \cite{nvidia}}          \\ \midrule
{\textbf{4. Data-driven {{3D simulation}}}}     & {\cmark}        & {\cmark}                              & {medium - high} & {medium - high}       & {implicit}  & {high}     & {low - medium}  & {E2E training / testing}           & {ViSTA 2.0 \cite{amini_vista_2021}, DriveGAN~\cite{Kim_2021_CVPR}} \\ \midrule
{\textbf{5. Mixed neural simulation}}       & {\cmark}        & {\cmark}                              & {medium - high} & {high}                & {mixed}     & {high}     & {medium - high} & {E2E training / testing}           & {-}                                                                                              \\ \bottomrule
\end{tabularx}
\label{table:levels}
\end{table*}
\end{flushleft}

% \vspace{-3em}  % remove weirdly large blank space before this paragraph

% As a basis for our comparisons, we provide the following definition of a simulation and apply it to specifically focus on simulations of the external environment and surroundings of AVs. We understand a simulation as the process of creating a simplified virtual replica of a subset of the real world in order to investigate its characteristics under different conditions in a controlled manner. It must allow for the exploration of hypothetical scenarios through varying its parameters.

\vspace{-2em}

We distinguish between the following levels or categories of simulation approaches. \\

%\subsection{Level 0: Log Replay}
%\label{sec:level0}

\textbf{Level 0: Log Replay.} Log replay is the most basic form of virtually reproducing real-world scenarios and involves to simply replay recorded driving data (e.g., streams of video data or point clouds, but possibly also CAN messages), without any way to change them, e.g., see~\cite{Lages2013}. Thus, it is not actually considered a type of simulation, but nevertheless widely used for testing and thus worth being mentioned in this taxonomy. Visual fidelity is high and only constrained by potential sensor imperfections. Diversity, on the other hand, is comparatively low, especially with respect to behavior. Recorded data will comprise corner case events (like accident scenarios or animals crossing a road) only very rarely. Scalability is very low, because new data must be collected each time in order to produce new variants of a scenario. Log replay allows developing and validating perception- and prediction algorithms. Due to its lack of reactivity, however, it does not enable for the exploration of hypothetical scenarios or testing different conditions.

%\subsection{Level 1: AR-enhanced Log Replay}
%\label{sec:level1}
\textbf{Level 1: AR-enhanced Log Replay.} AR-enhanced log replay builds upon plain log replay, but additionally allows to use augmented reality techniques to dynamically inject artificial objects into the world. This enables to add a certain degree of variability to otherwise static recordings and thus enables for better diversity. Recorded streams of exteroceptive sensor data, such as camera images or point clouds are modified retroactively to make synthetic 3D-modeled objects appear in the virtual scenes. This improves on scalability, because a single recording can be used for multiple different scenarios. However, the underlying data stream is still non-reactive, i.e., not influenced by actions taken in the virtual world, and thus not suitable for closed loop simulation, e.g., see~\cite{Zofka2014,Zofka2015}.

%\subsection{Level 2: Abstract Dynamics Simulation}
%\label{sec:level2}
\textbf{Level 2: Abstract Dynamics Simulation.} Shortcomings of levels 0 and 1 include the necessity for pre-recorded real-world logs and the fact that the simulated environments are not reactive to actions taken by the ego vehicle. Abstract dynamics simulations overcome these. These kinds of simulations can be run in abstract, simplified, entirely virtual environments and are closed-loop in that the environment can react to input signals dynamically. They usually involve a rather abstract, attributed object model, that only covers particular aspects, such as vehicle dynamics or behavior. Precise 3D geometry or texture information is usually not included, limiting content diversity to only a small set of parameters. As a consequence, however, these simulations allow for efficient execution even beyond real time and are easier to scale through sampling from the parameter space. Changing the fundamental structure (traffic participants, map layout, etc.) still involves manual work, however. Abstract dynamics simulations are well suited for simulating dynamic agents. Typical applications in AD include motion prediction, planning, and control. Level 2 simulations do not produce any type of sensor data (no camera, lidar, etc.) and thus are, by themselves, not suitable for perception tasks.

%\subsection{Level 3: Model-driven 3D Simulation}
%\label{sec:level3}
\textbf{Level 3: Model-driven 3D Simulation.} In addition to the functionality provided by level 2 simulations, model-driven simulators involve explicit physics- and object models, allowing for more advanced and comprehensive simulations, such as needed for developing and testing perception algorithms. They usually involve a 3D rendering engine and are conceptually similar to modern video games. Typical characteristics include the capability to provide multi-modal output (camera, lidar, radar, etc.), accurate physics simulation, accurate ground-truth labels and a fairly immersive, first-person-view experience for users. Their visual fidelity can vary largely between different implementations. These 3D engine-based simulators find application in end-to-end testing of driving stacks and are, along with level 2 simulations, most widely adopted today \cite{Kaur2021, zhang_cityflow_2019}. However, since 3D worlds, contents, and scenarios are mostly created manually by artists and domain experts, variability and scalability are still limited.

%\subsection{Level 4: Data-driven 3D Simulation}
%\label{sec:level4}
\textbf{Level 4: Data-driven 3D Simulation.} In order to overcome some of the model-driven simulations' limitations, a lot of research is put into data-driven approaches today. Various AD companies and research groups have demonstrated impressive results in this domain recently \cite{NVIDIA_WAD, Waabi2022, amini_vista_2021, Kim_2021_CVPR}. Data-driven simulators eliminate the need for explicit (3D-, behavior-, sensor-, etc.) models, but instead involve (generative) neural networks that learn to reconstruct or synthesize virtual worlds and behaviors from real-world data with little to no human supervision. They aim to produce photo-realistic outputs, are characterized through a small domain- and appearance gap and can be scaled easily, since not requiring a human in the loop. A drawback, however, is the often times limited control over the neural networks' outputs. While research is being done to allow for more human supervision (\hspace{-1sp}\cite{zhong2022guided, Kim_2021_CVPR}), parameters can usually only be tweaked implicitly through the input vectors. Moreover, most current approaches on data-driven simulators are capable of novel-viewpoint synthesis, but usually do not support synthesizing entirely new, unseen worlds.

%\subsection{Level 5: Mixed Neural Simulation}
%\label{sec:level5}
\textbf{Level 5: Mixed Neural Simulation.} Taking data-driven 3D simulations even one step further, mixed neural simulations are of particular interest for future research \cite{NVIDIA_WAD}. They address the previously mentioned limitations of level 4 and can be seen as a compound of multiple levels. Mixed neural simulations combine the data-driven nature of level 4 with the option to inject custom virtual objects from level 1 and fine-grained controllability from levels 2 and 3 and additionally complement support for full-synthetic environments. Generative capabilities are taken beyond only synthesizing novel viewpoints from given data and towards whole artificially generated worlds and behavior. Morevoer, neurally constructed, photo-realistic worlds can be integrated with arbitrary virtual content in a pick-and-place fashion for even higher diversity and flexibility. A key goal with this type of simulation is also to have precise control over the environment. However, to our knowledge, no simulators of this type are available today.

\begin{figure}
    \includegraphics[width=60mm]{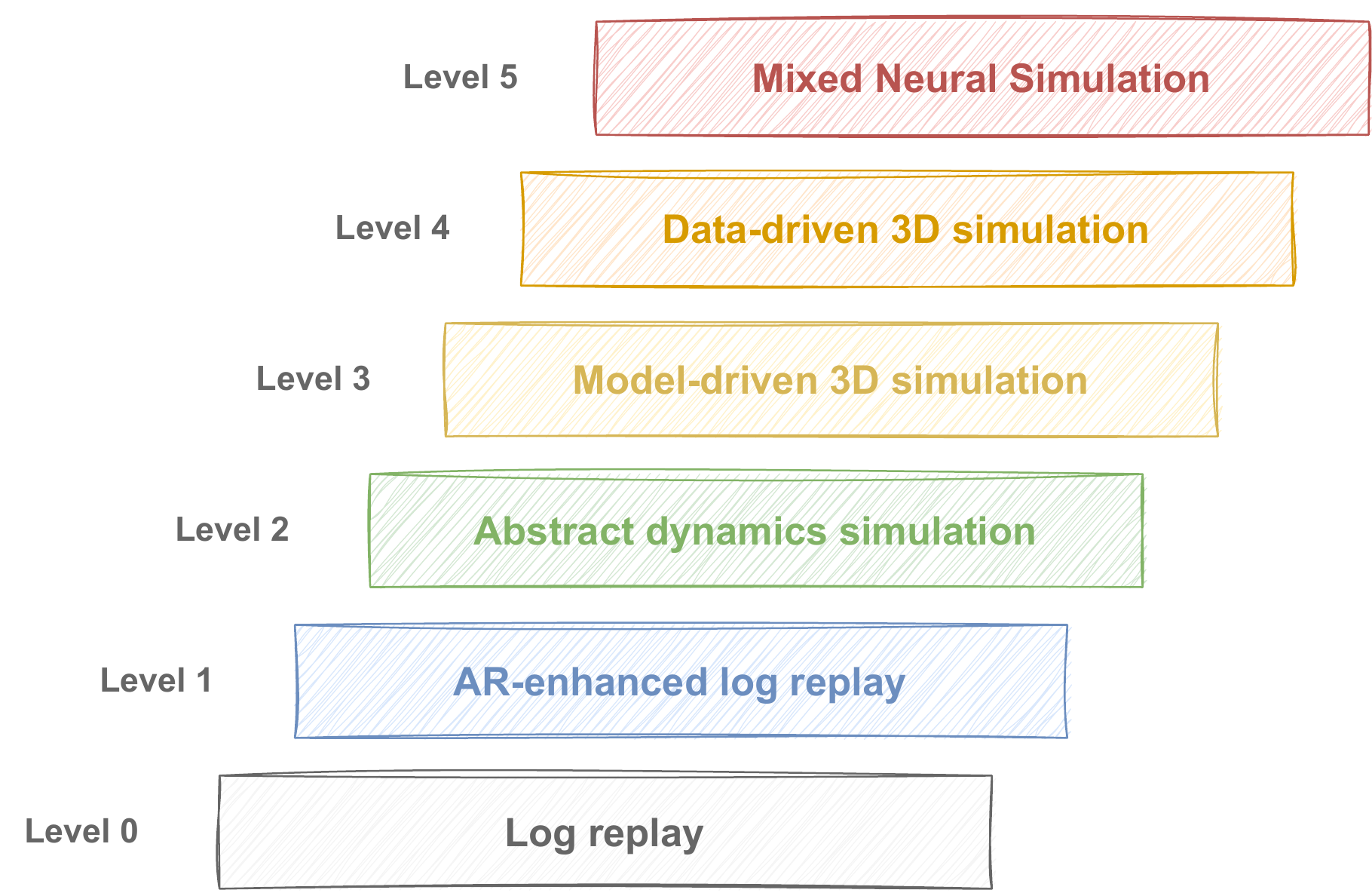}
    \centering
    \caption{Our proposed hierarchy of simulation levels}
    \label{fig:sim_levels}
\end{figure}

%\subsection{Conclusion}
\textbf{Conclusion.} We presented categories of simulation approaches for AD with a particular focus on simulation of an ego vehicle's external environment. While boundaries between these levels can be blurry sometimes, they nevertheless roughly build upon one another, allowing for increasingly complex and comprehensive, yet scalable, simulations. Higher levels typically allow to address challenges from different simulation aspects simultaneously, i.e., enable for high content-, behavior- and perception realism (see Sec.~\ref{sec:challenges_trends}) jointly within the same environment. The presented scheme helps to systematically classify modern simulation approaches in view of the field's latest advancements and is the first of its type. In the following, it serves as a guideline when investigating recent challenges and trends.

%% file: sections_paper/3_challenges_trends.tex
%%%%%%%%%%%%%%%%%%%%%%%%%%%%%%%%%%%%%%%%%%%
\section{Challenges and Trends}
\label{sec:challenges_trends}
%%%%%%%%%%%%%%%%%%%%%%%%%%%%%%%%%%%%%%%%%%%

\begin{figure*}
    \includegraphics[height=45mm]{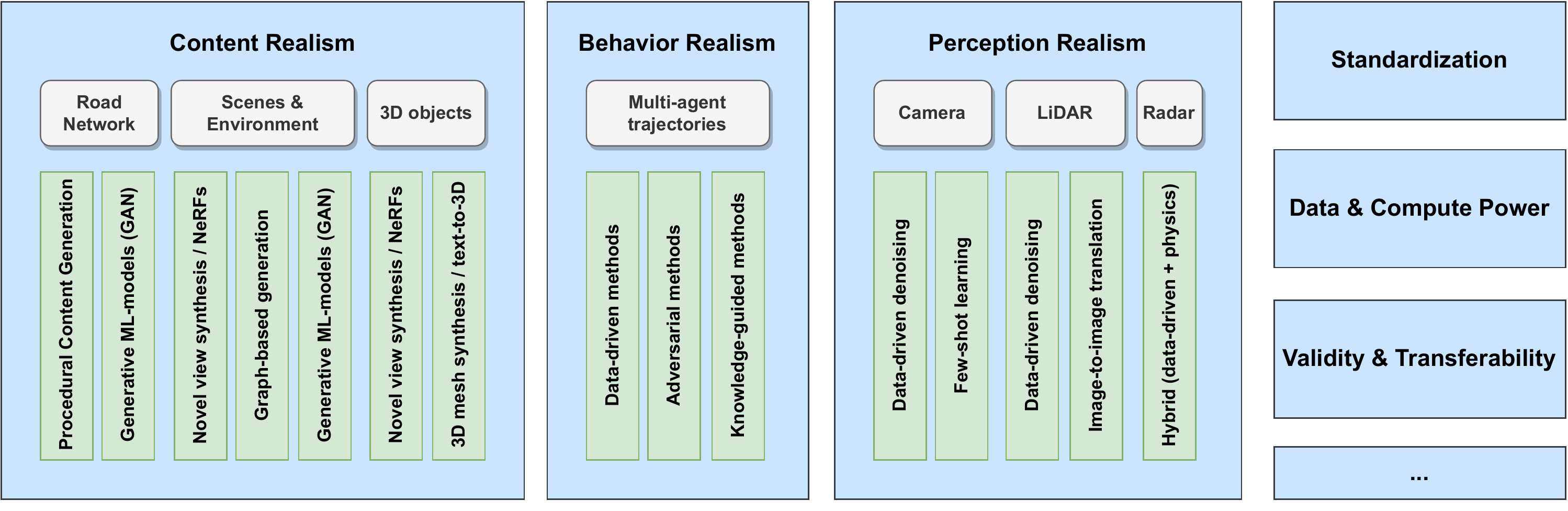}
    \centering
    \caption{Schematic overview of challenges \& trends in AD simulation}
    \label{fig:challenges_trends}
\end{figure*}

Picking up on the hierarchy of different simulation levels presented in Sec.~\ref{sec:levels}, we showcase current challenges and according trends (see Fig.~\ref{fig:challenges_trends}) on the path to reaching the upper levels in the following. We consider realism the overarching goal and biggest challenge on that path and define and decompose the term as follows.

Realism, in our understanding, describes the level of detail and accuracy of a depiction of reality. In the context of AD simulation, we distinguish between the following three different aspects of realism. The ultimate goal is to jointly maximize them, that is, minimize the gap between real world and simulation with respect to each of them. Firstly, \emph{content realism} is about accurately representing real-world objects (static and dynamic) and environments and their diversity. \emph{Behavior realism} covers dynamic aspects of real-world traffic scenes, such as non-static actors' motion characteristics. We exclude dynamics of the environment, such as changing weather effects here. Lastly, \emph{perception realism} is about replicating the appearance of the real world from the perspective of different sensors, e.g., to produce photo-realistic-looking RGB images or accurate lidar point clouds. Besides realism, a number of horizontal challenges, related to validity and transferability, data acquisition and -processing and standardization are presented in addition.

\subsection{Content Realism}
\label{sec:content_realism}
The first set of challenges focuses on realistic representations of a traffic scene's \emph{contents}. This includes its structure and topology itself, the road layout and -infrastructure and all static and dynamic objects surrounding the ego vehicle.

%\subsubsection{Road Network}
%\label{sec:road_network}
\textbf{Road Network.} An inherent part of driving simulations is the road network, usually represented as high-fidelity 3D maps. This includes the definition of road boundaries, driving lanes, traffic lights, stop signs and others. Traditionally, these maps were derived from the real world using 3D mapping techniques or created by hand in CAD. Neither approach is scalable to the requirements of automated virtual testing in AD. Therefore, procedural content generation (PCG) has emerged as a technique for synthesizing virtual environments~\cite{Freiknecht2017}. For example, PGDrive~\cite{Li2021} is a 3D simulator that procedurally generates randomized road networks based on a set of fundamental building blocks. More recently, generative deep-learning (DL) models, such as generative adversarial networks (GAN), were found to be applicable in this context as well ~\cite{Hartmann2017}. Lastly, Gisslén et al. proposed to employ adversarial reinforcement learning (RL) for PCG~\cite{gisslen_adversarial_2021}.

Each of these approaches eventually contributes to the scalable generation of diverse road maps and hence, in a broader sense, to higher diversity and realism of a traffic scene's content.

%\subsubsection{Scenes \& Environment}
%\label{sec:scenes_environment}
\textbf{Scenes \& Environment.} In addition to only the road network itself, many approaches attempt to construct entire scenes in 3D- or pixel-space, including static objects and the environment, such as buildings and vegetation. Traditionally, these were hand-crafted by artists, which is a tedious process that often times lacks variety. Many model-based 3D simulations additionally suffer from a severe lack of photo-realism \cite{amini_vista_2021}, but recent advances in computer graphics, such as with Unreal Engine~\cite{gamespot_unreal_2023}, promise to drastically alleviate this discrepancy.

SurfelGAN~\cite{Yang2020} is an approach to generate new scenes from data. Its authors propose the use of texture-mapped surfels (discs in 3D space) for reconstructing camera-recorded scenes in combination with a CycleGAN model, that additionally accounts for generation artifacts. SurfelGAN allows for both novel view point (NVP) synthesis and novel scene configuration, that is, perturbing objects in a scene to create variations.

Another popular recent trend is the use of Neural Radiance Fields (NeRFs) \cite{martinbrualla2020nerfw, Mildenhall2020} for NVP synthesis based on 2D imagery. Given a sparse set of input images of an object or a scene from different angles, NeRFs learn to project from a 5D-input vector (3D position and 2D viewing angle) to a pixel's precise color and depth information. This effectively yields a high-detail, 360° neural representation of the scene, including accurate lighting. Block-NeRF~\cite{Tancik2022} and BungeeNeRF~\cite{xiangli2022bungeenerf} extend the concept to large-scale scenes, such as entire housing blocks, through carefully combining multiple smaller NeRFs.

MIT ViSTA 2.0 \cite{amini_vista_2021} constitutes an entire open-source simulator that is solely data-driven. Given control inputs, it is capable of synthesizing lidar, RGB- \& event-camera observations for novel viewpoints, that are consistent with the virtual vehicle's kinematic model. This is achieved by complementing 2D RGB pixels with estimated depth information and subsequent projection into 3D space. As a result, ViSTA enables for developing end-to-end driving models with reinforcement- or guided policy-learning in simulation. Waabi World \cite{Waabi2022} seems to move in a similar direction, however, barely any details about its inner workings are publicly available. 

DriveGAN~\cite{Kim_2021_CVPR} addresses the typical problem of limited controllability over generated content. By decomposing the latent space into dedicated feature vectors for theme and content, the end-to-end differentiable, neural simulator allows to not only control background and weather of a scene, but also swap out objects at precise grid locations. While this level of control is still far from what model-based simulators provide, DriveGAN indicates an important direction for the future development of data-driven simulation. 

The previous approaches can be classified as data-driven (level 4), besides which there exist model-based approaches (level 3) as well. While the former usually generate sensor output directly, model-based approaches, in contrast, only yield more abstract structure and parameters and rely on a 3D engine for subsequent rendering. 

In this segment, Meta-Sim2~\cite{Devaranjan} proposed a way to generate realistically structured road scenes by utilizing an abstract graph representation of the scene. Using RL, their model is trained in an unsupervised fashion to sample rules from a probabilistic context-free grammar to iteratively build up a scene graph, while incorporating prior expert knowledge. Sim2SG ~\cite{prakash_self-supervised_2021} takes a similar direction. SceneGen~\cite{tan_scenegen_2021} is another approach towards traffic scene generation, based on ConvLSTM networks. These approaches yield diverse constellations of traffic agents in a scene and may also be used as an initialization for behavior models. Thus, they contribute to both content- and behavior realism.

%\subsubsection{3D objects}
%\label{sec:3d_objects}
\textbf{3D Objects.} Besides road network and general structure, another crucial part of simulated scenes are the contained objects, in particular, their precise geometry, texture, etc. These objects include cars, trucks, bicycles and pedestrians, but also buildings, trees, traffic signs and other road infrastructure. While a scene's fundamental structure defines what objects are located where, their precise properties are of importance, too. These objects must be both diverse and of authentic appearance. Most game-engine-based simulators only feature a limited set of mostly hand-crafted assets, which does not reflect the diversity of the real world and thus adds to a wider domain gap. Instead of having artists tediously design these objects, it is preferable to either extract them from data automatically or even let them be created synthetically. Various promising approaches have been developed either for 3D-aware novel view synthesis, such as with NeRFs, or actual 3D object generation. The latter are either represented in implicit (or neural) form, as point clouds or, more recently, as concrete 3D meshes.

Among the most promising recent advances in this regard is the GET3D model presented by Gao et al.~\cite{gao2022get3d}. They have developed an end-to-end trainable generative model that is capable of sampling entire 3D mesh models and according textures from latent space, solely trained on 2D imagery. While previous approaches either failed to capture geometric details, were limited in mesh topology or only produced implicit neural renderings, the outputs of GET3D are of high fidelity and can be used in 3D renderers directly. This helps to vastly increase object variety and build rich asset collections entirely automated. 

Further approaches are showcased by DreamFusion~\cite{poole2022dreamfusion} and Magic3D~\cite{lin2022magic3d}. Latest advances in diffusion models and NeRFs are leveraged to build text-to-3D generation methods, capable of synthesizing entire scenes from natural language input. In the context of autonomous driving, such can be used for both object- and whole road scene generation.

\textbf{Summary.} Self- or unsupervised deep generative models, especially NeRFs, find intensive application for both reconstruction and synthesis of simulation content and advance at a promising pace. Moreover, fully data-driven, synthesizing simulation environments rise in popularity over \mbox{3D-engine-}, model-based simulation systems. With regard to content realism, we see a trend from model-based simulation methods, involving  manually created road networks, scenes and 3D objects, towards deep-learning-facilitated, more automated approaches. Although recent works aim to address this, a key challenge, however, remains the limited degree of controllability with highly data-driven approaches.

\subsection{Behavior Realism}
\label{sec:behavior_realism}

In contrast to content realism, behavior realism is concerned with the dynamics of traffic scenarios. This includes viewing traffic flow from a macroscopic perspective, as well as considering individual agents' behaviors and precise driver attitudes or trajectories on a microscopic level. We focus on traffic participants' trajectories, but disregard behavior such as facial expressions or gesturing in the following. One of the biggest challenges is a lack in variety of traditional, parameterized models. Such fail to capture the full complexity of real-world traffic and often times miss out on the rare, unusual, long-tail events. Following the definitions of Ding et al.~\cite{Ding2022}, we distinguish between \emph{data-driven}, \emph{adversarial}, and \emph{knowledge-guided} methods.

%Please note that we do not break the topic of behavior realism down any further. Accordingly, other than previous and following sections, this section is now structured after different types of trends, as these are especially various for behavior. \\

%\subsubsection{Data-Driven Methods}
%\label{sec:data-driven_methods}

\textbf{Data-Driven Methods.} These methods involve learning traffic dynamics or agent behavior from examples to generate novel, but realistic new data. On the macroscopic side, Savva et al.~\cite{savva2019habitat} presented R-CTM, a scalable RNN-based approach for heterogeneous traffic flow simulation. TrafficSim~\cite{suo2021trafficsim}, on the other hand, is another approach for multi-agent motion prediction to overcome the limitations of traditional models. At least equally as interesting in the context of AD, however, are microscopic motion models.

To account for the temporal dimension of traffic scenarios, the vast majority of data-driven methods utilize sequence models, e.g. some sort of LSTM \cite{Deo2018, deo_multi-modal_2018, Lin2021, tan_scenegen_2021}. Explicitly incorporating the interactions and interdependence between nearby traffic agents turned out to be a notable success factors. According techniques include convolutional social pooling \cite{Deo2018} or the use of spatio-temporal attention \cite{Lin2021}. AgentFormer~\cite{yuan_agentformer_2021} is another recent attention-based approach. It utilizes the concept of transformer models to jointly incorporate both temporal- and social dimensions for trajectory prediction and has achieved significant improvement over the state-of-the-art. Given their success in other domains, we expect transformers to become much more important in the context of AD as well. Traditional sequence models like LSTMs or GRUs might get replaced to some extent in the future.

%\subsubsection{Adversarial Methods}
%\label{sec:adversarial_methods}
\textbf{Adversarial Methods.} Adversarial methods aim to particularly challenge the ego vehicle's planning. With adversarial reinforcement learning approaches, a generator network learns adversarial policies with the goal to interfere with the AV and try to make it fail. These approaches do not necessarily require naturalistic data and are very flexible. A particular challenge with adversarial methods is to produce trajectories that are difficult to cope with, yet not impossible. A challenge with adversarial RL particularly is to reward the agents properly in order to prevent them from just trying to collide with the AV, as this would result in unrealistic scenarios. For this reason, Wachi et al.~\cite{Wachi2019} divide the reward into an adversarial and a personal one. The adversarial reward is granted if the AV collides with any object in the virtual environment. The personal reward is granted to the agent if it maintains close to or reaches its personal destination, which is a certain location in the environment.

Other approaches use naturalistic data and modify the agents' trajectories to create critical scenarios. This can be achieved by formulating an optimization problem to make a scenario more critical, whereas criticality can be quantified through the use of different metrics \cite{9987969, westhofen2023criticality}, such as time-to-collision or drivable area. Klischat and Althoff ~\cite{Klischat2019} use evolutionary algorithms and try to minimize the drivable area of the AV by forcing the generators to adapt adversarial behavior. With NVIDIA STRIVE~\cite{rempe2022generating}, the model first learns a latent space representation of a traffic model before optimizing the agents' trajectories to challenge the AV. AdvSim~\cite{Wang_2021_CVPR} updates recorded lidar point clouds to match perturbed trajectories of surrounding vehicles to obtain scenarios that are safety-critical for a full autonomy stack. 

Like the data-driven approaches presented earlier, adversarial approaches contribute to more diverse, yet realistic behavior. They are comparatively novel and differ in being specifically purposed to provoke especially demanding situations. Often times, they also require less to no example data for doing so.

%\subsubsection{Knowledge-Guided Methods}
%\label{sec:knowledge-guided_methods}
\textbf{Knowledge-Guided Methods.} In a broad sense, we consider knowledge-guided methods, also referred to as Informed ML ~\cite{von2021informed}, as ML methods that incorporate prior expert knowledge to accelerate learning and yield more realistic results quicker. A conditional diffusion model for controllable realistic traffic was developed by Zhong et al. \cite{zhong2022guided}. Here, a diffusion model for trajectory generation is combined with Signal Temporal Logic (STL). STL can be used to define rules such as obeying regulations or avoiding collisions and thus make the generated trajectories fulfill certain criteria. While these methods are not conceptually new, they had not found substantial application in AD so far.

\textbf{Summary.} In behavior realism, there is a trends towards increased use of ML. Data-driven methods mainly utilize sequence models and generate trajectories depending on surrounding agents and the environment. Those methods are well suited for realistic standard behavior as occurs in the training data. Adversarial methods, on the other hand, try to challenge the AV and have the explicit goal to produce critical scenarios. This is particularly important for V\&V of the AV. Knowledge-guided methods incorporate expert knowledge to gain better control over generated trajectories.

\subsection{Perception Realism}
\label{sec:perception_realism}
Perception realism, according to our understanding, is about accurately modelling the specific characteristics and noise distributions of different types of (perception-related) sensors. Most prominent for AV development today are camera, lidar and radar sensors, the simulation of each of which faces individual challenges.

%\subsubsection{Camera}
%\label{sec:camera}
\textbf{Camera.} Cameras are semantically the richest type of perception sensor and common among many applications. However, their accurate simulation still remains challenging. Different camera types and models suffer from different sorts of noise, such as under- and overexposure or distortion due to lens geometry or shutter speed. To minimize the real-to-sim gap, models used for simulation must account for these sensor-specific characteristic and / or generalize from them. Traditionally, statistical noise models were used, however, such cannot fully account for real-world complexity. 

In recent research, DL methods are used to infer noise models from data. Chang et al.~\cite{chang2020learning} presented a generative approach for image denoising which supports multiple cameras simultaneously. The authors emphasize their future ambitions to extend the model to support single- or few-shot learning in order to reduce data requirements for adapting new cameras. Similarly, FFDNet~\cite{8365806} was developed as a particularly fast denoising model, based on CNNs and with support for different noise levels.

%\subsubsection{LiDAR}
%\label{sec:lidar}
\textbf{Lidar.} Traditionally, simulation of point clouds is mostly based on depth maps and raycasting~\cite{yu2022autonomous}. While constructing lidar points geometrically is fairly well understood, the challenge lies in accurately simulating intensity, attenuation and raydrop. Raytracing techniques enable to account for the intensity, however, are computationally expensive and necessitate explicit, high-detail information about the objects' geometry and texture.

Recent approaches to make sensor simulation more effective are based on data-driven methods, which apply ML to learn to reproduce sensor data from recordings. Marcus et al.~\cite{Marcus2022} translate point prediction into an image-to-image translation problem to generate 3D lidar points from 2D images. A different approach is pursued with LiDARSim~\cite{Manivasagam_2020_CVPR}, where the authors first reconstruct a 3D scene in simulation, then use traditional raycasting to get point clouds and eventually apply a deep neural network to insert more realistic patterns in the form of deviations and noise. Another example in this realm is RINet~\cite{microsoft_research_data-driven_2022}, which uses supervised learning to obtain a model for raydrop and intensity simulation from RGB images.

%\subsubsection{Radar}
%\label{sec:radar}
\textbf{Radar.} Radars are seldom used directly for scene understanding, but rather to support camera- or lidar perception \cite{8911477}. While featuring superior performance under adverse weather conditions and for the purpose of measuring relative speed, radar sensors are especially difficult to model. That is due to physical effects and phenomena such as multi-path reflections, interference, ghost objects or attenuation~\cite{Wheeler2017}. Moreover, radar data are more sparse and stochastic compared to lidar. While the stochasticity aspect can be addressed by including detection probability as an additional random component to the simulation \cite{Ngo2021}, the impracticability of complete reconstruction of radar wave remains. A recent survey presents different schemata for classifying radar models by fidelity \cite{s22155693}. One can distinguish between idealized (aka. black-box), physics-based (aka. white-box) and hybrid (aka. grey-box) models, whereas the latter two are of most relevance in AD.

Simulating radar data, especially including realistic sensor noise, is challenging. However, even though this is an ongoing research topic (cf. \cite{10.1007/978-981-16-0739-4_32, mathworks_radar_nodate, peng_radarsimpy_2023}), we could not observe any clear trends in this regard.

\textbf{Summary.} In sensor simulation, a generally observable trend is to leverage real-world data in learned models to boost the capabilities of classical, statistical- or physics-based models. Few-shot learning appears to emerge as a promising way to account for covering the great variety of different sensor models and improve perception realism.

%%%%%%%%%%%%%%%%%%%%%%%%%%%%%%%%%%%%%%%%%%
\subsection{Horizontal Challenges}
\label{sec:horizontal}
%%%%%%%%%%%%%%%%%%%%%%%%%%%%%%%%%%%%%%%%%%

Previous sections focused on challenges and trends around realism, as such is the primary goal with every simulation. However, there are a number of cross-cutting, horizontal challenges as well, that span through all types and levels of simulation.

\textbf{Standardization.} As more AD simulation tools emerge on the market, the need for a standardized format for scenario description is growing in order to facilitate transfer and exchange between platforms. A current approach to tackle this is ASAM's\footnote{Association for Standardization of Automation and Measuring Systems} family of OpenX standards, including OpenSCENARIO~\cite{association_for_standardization_of_automation_and_measuring_systems_openscenario_nodate} for describing the dynamics of traffic situations, OpenDRIVE for maps, and OpenCRG ~\cite{association_for_standardization_of_automation_and_measuring_systems_opencrg_nodate} for road surfaces, complemented by OpenMaterial ~\cite{friedman_openmaterial_2023}, developed by BMW, for description of material properties. In addition, the Open Simulation Interface (OSI) ~\cite{osi} exists to support data exchange between different simulation models, while Eclipse Cloe ~\cite{eclipse_foundation_eclipse_2020} contributes a middleware layer to abstract from specific simulation environments and thus speed up development. Having big industry players collaborate on open standards is a welcomed trend for AV development in general. However, despite these standardization efforts, other simulation data formats, such as GeoScenario \cite{Queiroz2019} by University of Waterloo, TUM CommonRoad scenarios \cite{Althoff2017}, Traffic Charts and Scenic \cite{Fremont2022} by UC Berkeley are still being developed in parallel.

\textbf{Data \& Compute Power.} Constructing high-fidelity simulations requires vast amounts of data to ensure that the simulated environment is representative of the world it is supposed to model. In AD, these data may originate from real-world test drives, from a fleet of sensor-equipped consumer cars or from road-side measuring infrastructure. They are generally hard to gather at scale for organizations other than big industry player. We hope to see a growing trend towards open-data initiatives in the future. Moreover, compute power becomes an increasingly important factor, while its limited availability is often times a bottleneck to scalability. The development of specialized, high-efficiency machine-learning hardware may help in this regard.

\textbf{Validity \& Transferability.} Another general challenge with simulations is to quantify their real-world validity. On the one hand, one commonly seeks to minimize the real-to-sim gap. Closing the sim-to-real gap, however, is equally as important. For a model developed and tested virtually, it is crucial to ensure its applicability to the real world. The most important question is around how to measure if a simulation is \emph{good enough} to be used as a replacement for real-world testing and training. This also relates to AI safety and is a research field of tremendous popularity today.

%% file: sections_paper/4_conclusion.tex
%%%%%%%%%%%%%%%%%%%%%%%%%%%%%%%%%%%%%%%%%%
\section{Conclusion}
\label{sec:conclusion}
%%%%%%%%%%%%%%%%%%%%%%%%%%%%%%%%%%%%%%%%%%
In this work, we first proposed a novel classification scheme to systematically compare AD simulation approaches, that takes latest advancements on the topics of data-driven and neural simulators into account. We then outlined current challenges in that domain and presented recent research trends aiming to address them. We specifically focused on the content-, behavior- and perception realism and also touched upon a few cross-cutting challenges.

With regard to content realism, self-supervised generative models, such as NeRFs, enable to synthesize virtual 3D worlds with little to no hand-crafted data required. Behavior realism benefits from recent advances in deep learning, such as elaborate sequence models or transformers, but, more recently, also from adversarial models trained with RL. Deep learning models have gained ground for perception realism as well and are often times used to augment classical models with data-informed insights. Few-shot learning emerges as a promising way to cover a greater variety of sensor characteristics. A general challenge in AD simulation is a lack of standardization, especially in terms of data formats. However, industry seems to slowly converge towards common standards on that end. Many questions around validity \& transferability of simulations as well as data- and compute requirements still remain subject to ongoing research.

Across all aspects of a simulation, data-driven methods gain popularity over model-based approaches and promise to overcome many of their limitations. Given these trends, simulation environments are being rapidly towards levels 4 and 5, that is, comprehensive data-driven- and mixed neural simulations. We expect future research to continue in this direction with a fast pace, enabled through the application of (generative) deep learning across all areas of AD simulation. Heading in this direction, we identified a number of important future research questions, including:

\begin{enumerate}
    \itemsep-0.5em 
    \item How can data-driven methods and their parameters be made better understandable and configurable in order to leverage expert knowledge?
    \item How can quality, validity and transferability of simulations, and data-driven approaches in particular, be quantified?
    \item How can we extract or generate training data from the real world or in simulation in a large-scale manner and assess its relevance for different ODDs?
    \item How can AV development benefit from incorporating subtle, high-detail phenomena, such as gestures and facial expressions, in simulations?
\end{enumerate}

With respect to limitations and future work, it would be of interest to conduct a larger and more systematic follow-up literature review, that goes into greater detail and also covers aspects that were intentionally disregarded in this work. Additionally, as many simulators are not available in an open-source fashion, gaining knowledge about these is important for more detailed comparisons in the future.

%% file: sections_paper/5_acknowledgment.tex
\section{Acknowledgment}
\label{sec:acknowledgment}
The research leading to these results is funded by the German Federal Ministry for Economic Affairs and Climate Action within the "VVMethoden" (19A19002A) and "SofDCar" (19S21002) projects. The authors would like to thank both consortia for the successful cooperation.